\documentclass[sigconf]{acmart}

\usepackage{booktabs} % For formal tables
\usepackage{caption}
\usepackage{subcaption}
\usepackage{enumitem}
\setlist{leftmargin=*}
\setcopyright{acmcopyright}
\acmDOI{xx.xxx/xxx_x}

\acmISBN{978-1-4503-8713-2/22/04}

\acmConference[SAC'22]{ACM SAC Conference}{April 25 –April 29, 2022}{Brno, Czech Republic}
\acmYear{2022}
\copyrightyear{2022}

\acmArticle{4}
\acmPrice{15.00}

\begin{document}
\title{An Edge Map based Ensemble Solution to Detect Water Level in Stream}
% \titlenote{Produces the permission block, and
% \subtitle{Extended Abstract}
% \subtitlenote{The full version of the author's guide is available as
%   \texttt{acmart.pdf} document}
  
\renewcommand{\shorttitle}{SIG Proceedings Paper in LaTeX Format}

\author{Pratool Bharti}
\affiliation{%
  \institution{Northern Illinois University}
}
\email{pbharti@niu.edu}

\author{Priyanjani Chandra}

\affiliation{%
  \institution{Northern Illinois University}
}
\email{z1864520@students.niu.edu}

\author{Michael E. Papka}

\affiliation{%
  \institution{Northern Illinois University} 
  \institution{Argonne National Laboratory}
}
\email{papka@niu.edu}
\author{David Koop}

\affiliation{%
  \institution{Northern Illinois University}
}
\email{dakoop@niu.edu}

\begin{abstract}
Flooding is one of the most dangerous weather events today.  Between $2015-2019$, on average, flooding has caused more than $130$ deaths every year in the USA alone \cite{c1}. The devastating nature of flood necessitates the continuous monitoring of water level in the rivers and streams to detect the incoming flood. In this work, we have designed and implemented an efficient vision-based ensemble solution to continuously detect the water level in the creek. Our solution adapts template matching algorithm to find the region of interest by leveraging edge maps, and combines two parallel approach to identify the water level. While first approach fits a linear regression model in edge map to identify the water line, second approach uses a split sliding window to compute the sum of squared difference in pixel intensities to find the water surface. We evaluated the proposed system on $4306$ images collected between $3$rd October and $18$th December in 2019 with the frequency of $1$ image in every $10$ minutes. The system exhibited low error rate as it achieved $4.8$, $3.1\%$ and $0.92$ scores for MAE, MAPE and $R^2$ evaluation metrics, respectively. We believe the proposed solution is very practical as it is pervasive, accurate, doesn't require installation of any additional infrastructure in the water body and can be easily adapted to other locations.
\end{abstract}




\maketitle

\section{Introduction}

% Points to describe here
% 1. Flash flood in urban settings 
% 2. Proliferation of camera 
% 3. Privacy is paramount in multipurpose camera
% 4. Model should be able to say ``decision not available" when not confident
% 5. Accuracy should be high
% 6. Light weight algorithm that can run on resource scarce device

Flooding is one of the deadliest weather-related hazards in US \cite{usdc}. Deaths due to flooding are often associated with excessive rainfall that leads to flash flooding.  Flash floods can occur within a few minutes or hours of excessive rainfall in urban areas with a poor drainage system. With an alarming rise in global temperature, many parts of the world are suffering from recurrent floods and excessive rainfall \cite{unep}. The National Oceanic and Atmospheric Administration (NOAA) have reported that record-breaking rainfall in Louisiana in 2016 were at least $40$\% more likely and $10$\% more intense due to the rapid climate change \cite{c6}. Estimating revenue loss, floods have caused approximately \$$1.6$ billion worth of property and crop damage across USA in $2018$ alone \cite{c3}. A recent study from the SHELDUS  database \cite{c4} estimate that a total of \$$107.8$ billion in direct property damage incurred in urban areas due to flooding affecting $20,141$ urban counties between $1960$ and $2016$.

Needless to say, an early and effective flood warning system can potentially save hundreds of lives and allow individuals and communities enough time to protect their valuables. These systems can frequently monitor the change in water level in rivers and streams in flood-prone areas and send public alerts when an incoming flood is detected. In US, United States Geological Survey (USGS) maintains a network of over $8,200$ streamflow-gaging stations throughout the country to measure the water level in the rivers and streams \cite{c7}. In one of the common approach, they install a stilling well in the river bank.  The well is connected with the river through underwater pipes to keep the water level in the well at the same height as in the river. Subsequently, water level in the stilling well is measured using a float or a pressure, optic, or acoustic sensor. This process is repeated in every 15 minutes to capture the electronic record of water level. Although this approach provides accurate readings, it requires to excavate a stilling well by the river which is costly to install and maintain. The National Weather Service (NWS) sends flood warnings to local communities in US by detecting heavy rainfall associated with flash floods. They primarily use satellite, lightning observing systems, radar, and rain gauges to measure the intensity of the rain \cite{floods101}. While excessive rainfall is one of the major cause of flash floods, it is not a direct measurement of flood occurrence and may lead to inaccurate predictions. However, directly gauging the water level in the stream provides prompt assessment of flood risk and offer more accurate prediction.

\begin{figure*}[!h]
    \centering
    \begin{subfigure}[h]{0.49\linewidth}
        \includegraphics[width=\linewidth]{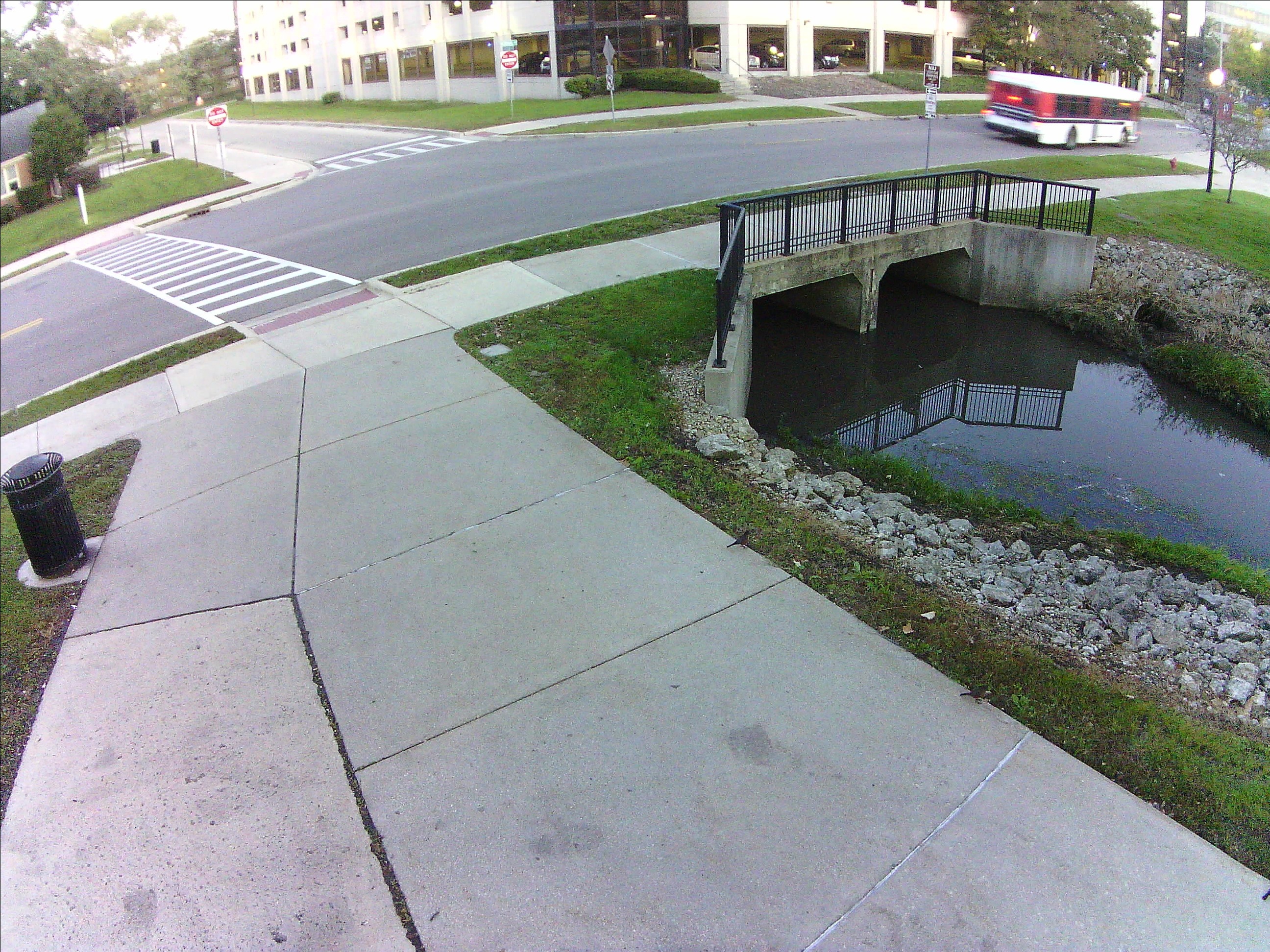}
        \caption{A sample color image from camera}
    \end{subfigure}
    \begin{subfigure}[h]{0.49\linewidth}
        \includegraphics[width=\linewidth]{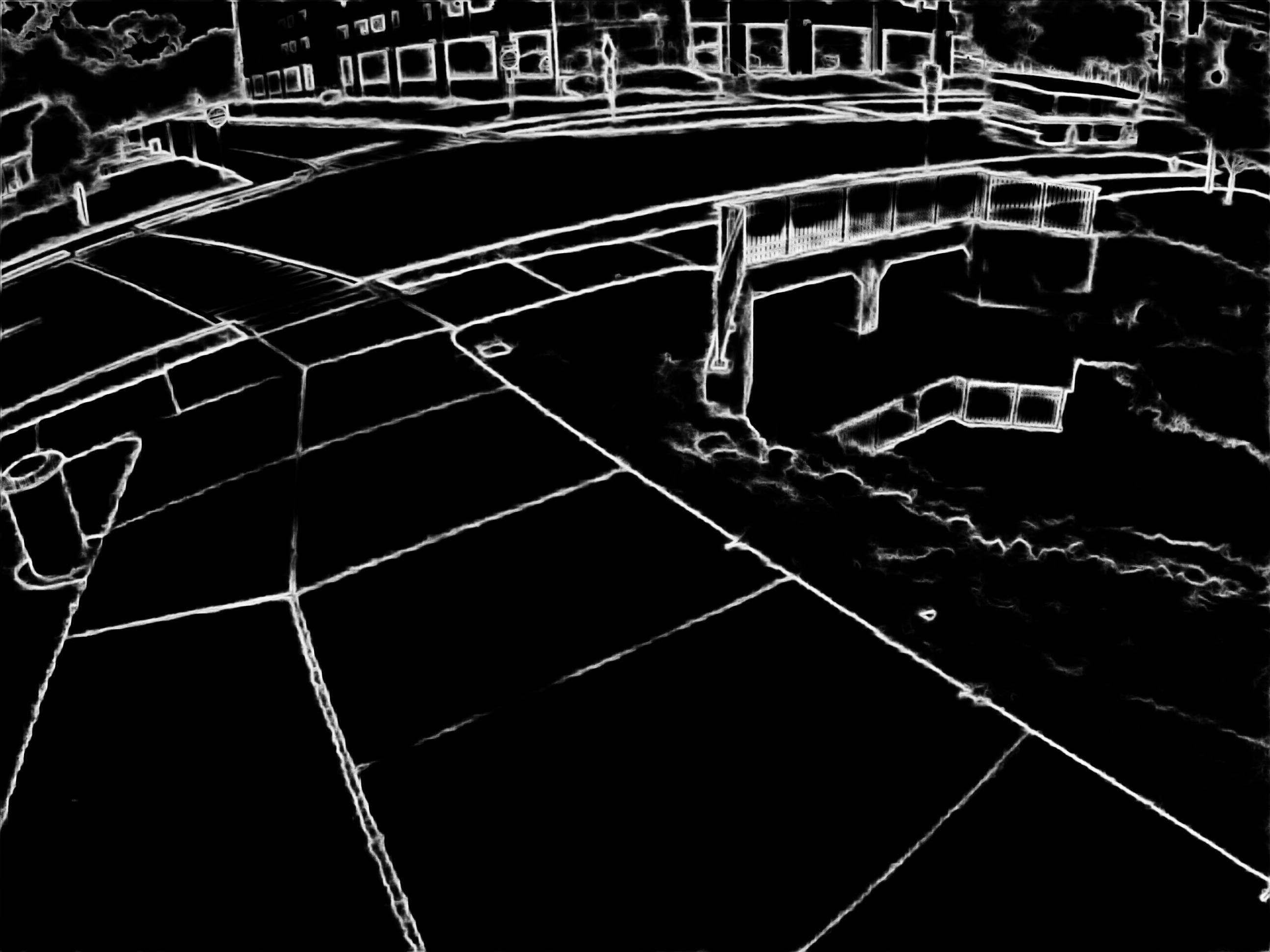}
        \caption{Edge map image after applying HED}
    \end{subfigure}
    \caption{A color and edge map image}
    \label{fig:original_and_hed}
\end{figure*}

The limitations of existing flood detecting systems necessitates a solution that constantly monitors the water level in the streams with high precision. The most adequate solution should be automated, cost effective and accurate. With the proliferation of low-cost vision cameras and advancements in image processing techniques, an image-based water level detection system can be pervasive and cost-effective. Additionally, a single camera feed can be used to find solutions in different contexts. For example, a camera installed on an intersection can be useful to design automated solutions to detect traffic jams, accidents, over-speeding cars, finding missing person, and more. In this realm, we propose an efficient vision-based system to automatically detect the water level in the stream using a single image. The proposed solution leverages images from a camera installed in a university campus that oversees the traffic as well as water in the creek (as shown in Fig. \ref{fig:original_and_hed}). We believe our solution can be effective and widely acceptable as it comprises of three important components - high accuracy, trustworthiness, and efficiency (lightweight). To accomplish them, our main contributions in this work are as follows -

\begin{itemize}
%   \item To preserve the privacy of pedestrians, the proposed system doesn't use raw color images instead it works with edge map for recognizing water surface (as shown in Fig. \ref{fig:original_and_hed})
  \item To produce accurate prediction, the proposed ensemble solution combines two different methods (linear regression based and split sliding window based) to find the water level. The system generates the outputs only when both approaches converge on a single response.
  \item To generate fast response, it avoids training  complicated machine learning models instead it relies on pretrained holistically-nested edge detection (HED) \cite{xie2015holistically} and normalized cross-correlation template matching techniques to find the region of interest in the image.
  \item To make the solution trustworthy, the proposed system provides the water level readings only when it finds the quality of input image satisfactory to avoid any misleading responses caused by bad inputs. Additionally, it generates no response when both proposed approaches do not converge on same response. 

\end{itemize}

In the next few sections, we discuss the previous related research works followed by details on data collection, approach and results.
\section{Related works}
In recent years, researchers have proposed multitude of solutions to estimate the change in water level in a stream. Majority of these solutions are leveraging ultrasonic and infrared sensors \cite{mousa2016flash} \cite{malek2020real}, synthetic aperture radar \cite{martinis2015comparing}, rainfall and pressure sensors \cite{basha2008model}, and vision camera \cite{lai2007real}. Except vision based solutions, almost every other approach requires to maintain a sophisticated infrastructure that comes with a high cost, complex to install and difficult to use. With the recent proliferation of vision camera, image and video based solutions have become low-cost, pervasive and straight forward to install and use. In this section, we discuss the vision based research works in detecting water level in a stream. For simplicity, we have divided them into two categories - vision camera with staff gauge or other markers, and without staff gauges.

In first category, Jafari et al.  \cite{jafari2021real} have trained a deep neural network based AI model to detect the water body and bridge pier through image segmentation. The model computes the change in the water level based on detected change in bridge pier height (the visible height of bridge pier decreases as it submerges in water). Although the proposed solution has relatively low error rate, training a deep neural network is challenging as it requires tagging of large amount of images to prepare the training dataset. More importantly, these solutions are not transferable from one site to another due to the high variance in images, hence, the whole process would require to repeat from the scratch. In another interesting work \cite{wang2018integrated}, Ridolfi and Manciola have used images from drone camera to measure the water level at a dam site. Proposed solution leverages Canny edge detector algorithm \cite{canny1986computational} to find the edge of water level which is highly sensitive to the intensity of light in the image \cite{arandiga2010edge} and doesn't work well in poor light conditions.

Among studies with staff gauge installed in water body, Kuo \& Tai  \cite{kuo2020implementation} have proposed an image processing technique to detect the water level in a drainage channel by reading the measurements from an onsite staff gauge. They have used an inverse perspective mapping technique to fix the perspective distortion in the image induced by non-orthogonal camera position. In another similar work, Zhen et al. \cite{zhen2019visual} have employed fusion of gray and edge features extracted to improve the appearance of staff gauge in the image sharper. The proposed solution has demonstrated improved performance under poor light conditions. In \cite{lin2018automatic}, Lin et al. have used single camera image to detect the water level with the help of water gauge. They have employed photogrammetric technique that tracks the unwanted camera movements and minimizes its effects. In \cite{lin2013applications}, Lin et al. have used standard image
processing techniques for image binarization and subsequently recognizes digits from water gauge to capture water level readings. While all of these works show significant improvement in performance by enhancing the image quality, they are exceedingly relying on staff gauge to measure the water level.

\section{Data Collection}
\begin{figure}[h!]
    \begin{subfigure}{0.95\linewidth}
        \centering
        \includegraphics[width=\linewidth]{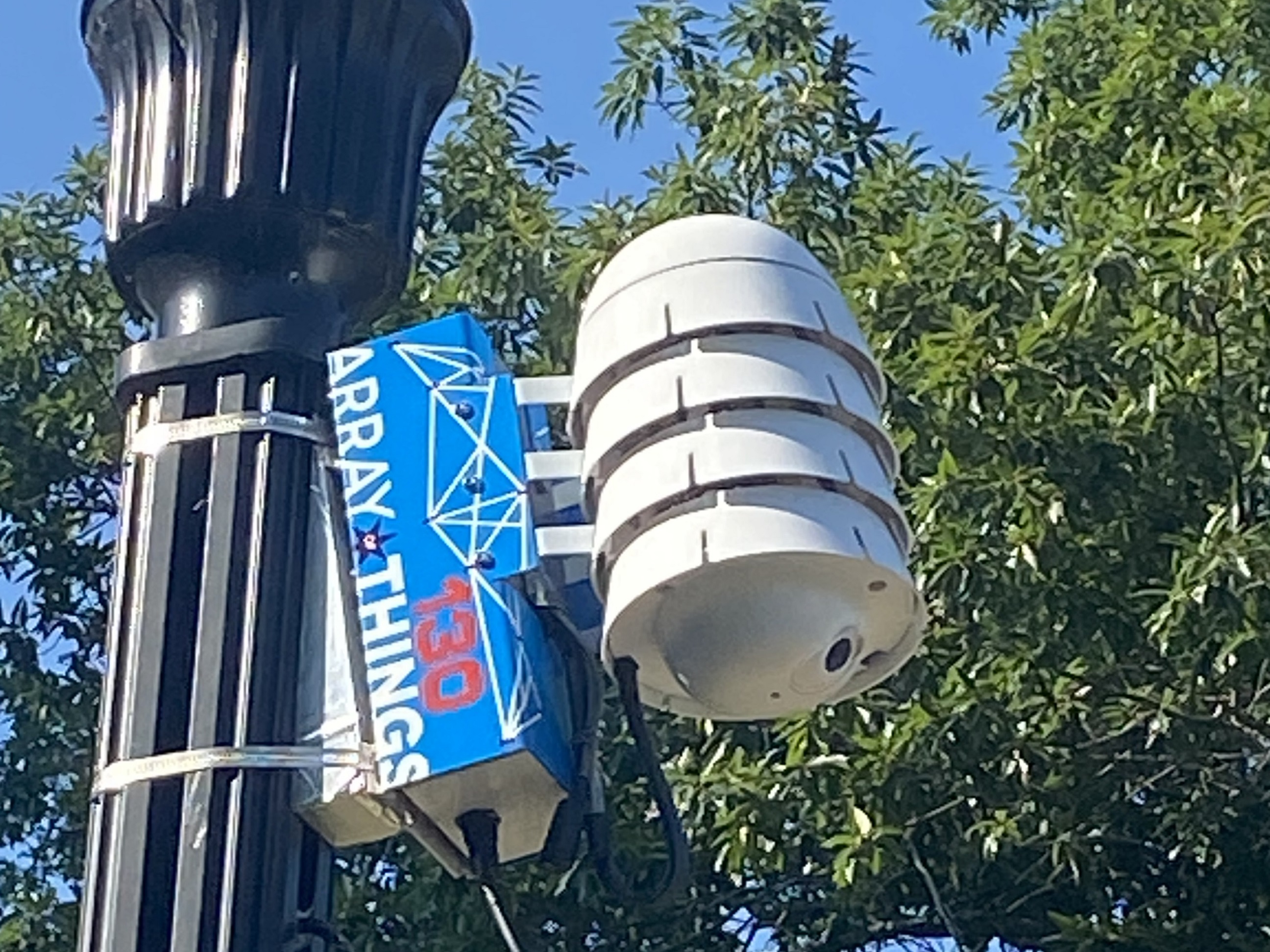}
    \end{subfigure}

    \caption{AoT node installed on a light pole}
    \label{fig:aot_and_image}
\end{figure}

We installed an Array of Things (AoT) node~\cite{beckman2016waggle} on a light pole in a university campus that oversees water in the creek along with traffic on the street (as shown in Fig. \ref{fig:aot_and_image}). 
% The data, resources, and infrastructure required to conduct the research are provided by the data, devices, and interaction Lab (ddiLab) at Northern Illinois University (NIU).   
AoT is an experimental system to study urban life and the environment by collecting real-time data using a sensor platform with edge computing capability. AoT sensor suite includes two $5$ MP cameras, environmental, air quality, and light and infrared sensors to monitor street conditions, weather conditions, air quality, sound, cloud cover, and sunlight intensity. The node is embodied with two Linux-based computers. While one computer is equipped with an Amlogic quad-core ARM Cortex-A5 processor to manage the sensory data and network stack, other computer contains Samsung Exynos5422 CPU with $4$ A15 cores, $4$ A7 cores, and a Mali-T628 MP6 (GPU) that supports OpenCV~\cite{bradski2000opencv} and Berkely Caffe~\cite{jia2014caffe} software packages. It also consists of a microcontroller board designed for managing power, programming of computer boards, and provides resilience to the system. 

\begin{figure}[!h]
    \centering
    \includegraphics[width=\linewidth]{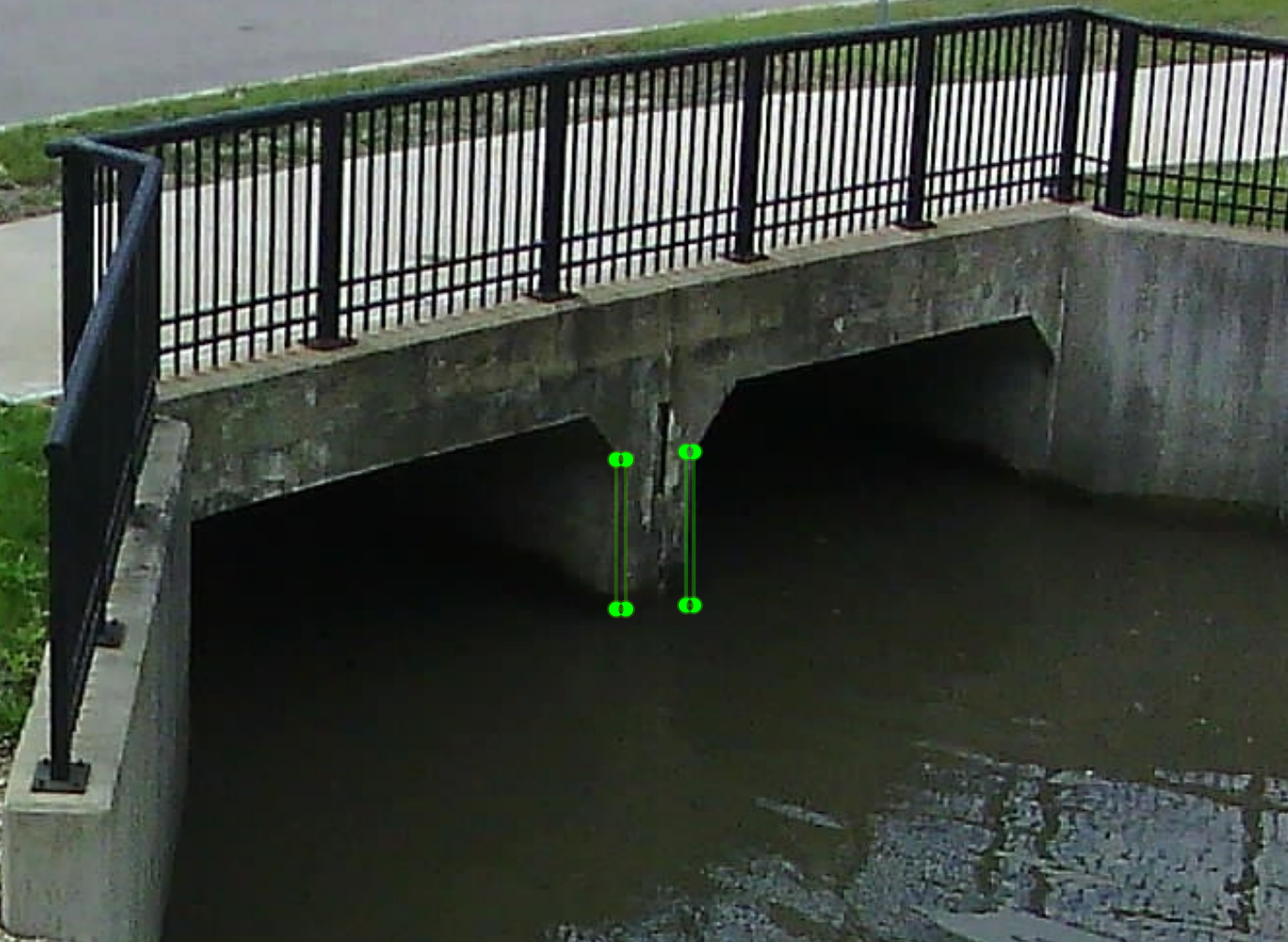}
    \caption{Annotation of water level using LabelImg tool}
    \label{fig:gt_annotation}
\end{figure}

The node captured $2,583,646$ images of creek with the rate of $1$ image per second between $3$rd October and $18$th December in $2019$. The resolution and size of each image are $96$ dpi and $2592\times1944$, respectively. Due to  technical issues, images for few days were missed during this period. In this study, we employed total $4306$ images with $1$ image every $10$ minutes to determine the water level in the creek. The reason to pick $10$ minutes delay was solely driven by the manual effort required to tag the water level in each image to prepare the ground truths. We leveraged LabelImg tool \cite{tzutalin_labelImg} to manually annotate each image at $2$ places to minimize any human error (as shown in Fig. \ref{fig:gt_annotation}). Finally, the distance of water surface from a fixed location of the bridge was measured in pixels to compare against the prediction from the proposed system. 

\begin{figure*}[h!]
    \centering
    \includegraphics[width=\linewidth]{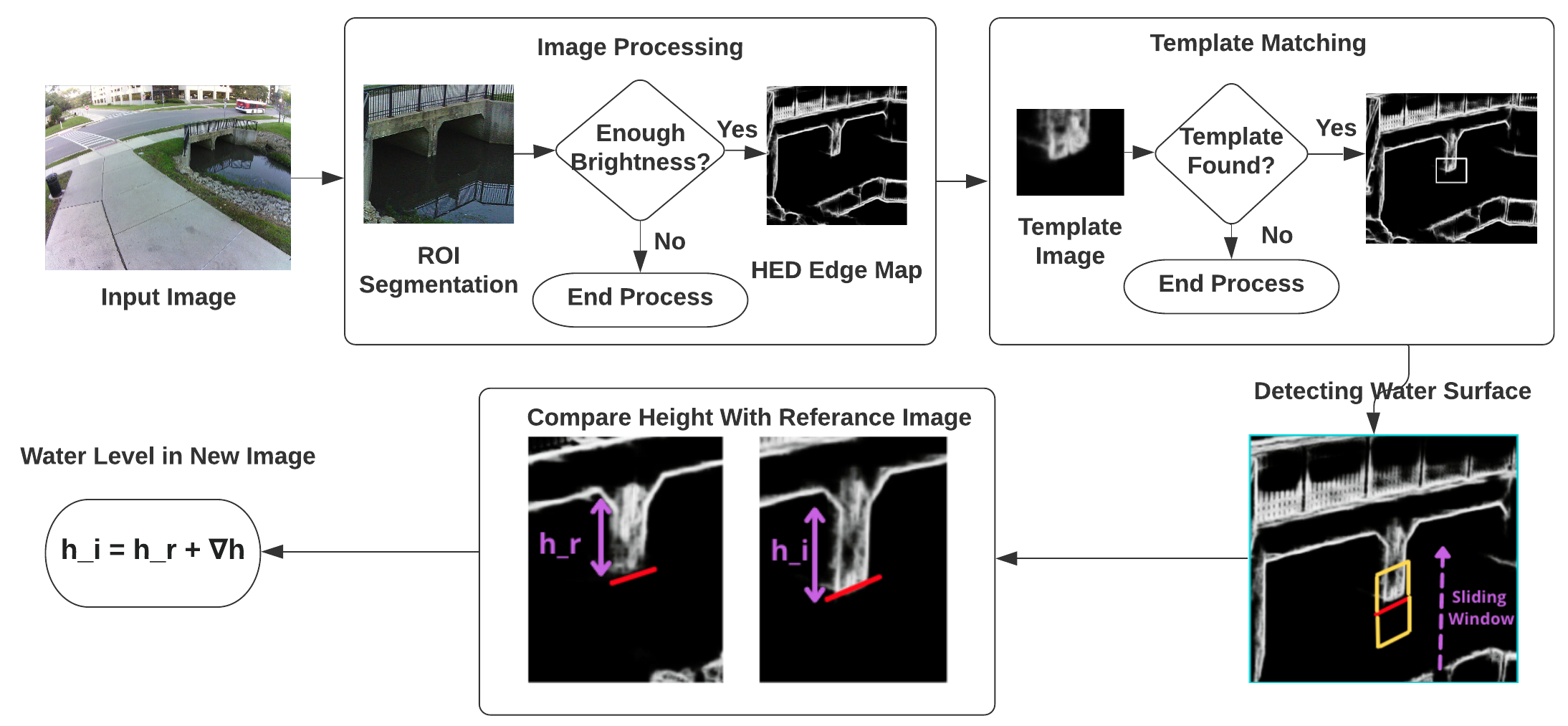}
    \caption{The System Workflow}
    \label{fig:flow_diagram}
\end{figure*}

\section{System Overview}
% The literature review of related works helped in finding various existing solutions and their problems in finding the change in water level of a water body. It also helped me in understanding the process and analyzing the steps required to complete my research. Analyzing image data and perceiving approaches that work on the available data are key tasks in the research. Utilizing Computer Vision and Machine Learning concepts for detecting the water level is a forethought to achieve the goal.
As discussed in related works, majority of the previous related works either rely on a staff gauge installed in the water body or train a complex AI algorithm that takes large annotated training dataset but doesn't generalize well on other locations. Despite several advantages, image based water level detection is a challenging problem where majority of challenges arise due to the noise or variance in the image induced by poor light and weather conditions, occlusion, shadow and floating debris or waves in the water. In these conditions, a vision-based system can easily make mistakes and provide wrong responses causing distrust in the system. A good solution must know its limitation and alert the users when not confident about their responses. To make the proposed system trustworthy, it screens the quality of input image using several checks and only response when is in good confidence. Furthermore, it generates the final response only when the two parallel approach converge on single response. The proposed system employs lightweight image processing techniques that doesn't require any additional infrastructure or large amount of training data to accurately measure the water level in the stream. 

Here, we briefly describe the workflow of our system (shown in Fig. \ref{fig:flow_diagram})). The complete pipeline is consists of $4$ steps. First, image pre-processing where the system segments the ROI from full image and reduces the noise by passing it through a low pass filter. Second, it transforms segmented ROI image from color to edge map to highlight the water surface. It leverages holistically-nested edge detection algorithm \cite{xie2015holistically} to generate the edge maps. Third, a template matching (TM) algorithm is adapted to further reduce the ROI to a small region where water surface meets the bridge pier. TM acts as another check to measure the image quality. Fourth, it uses two parallel approach to detect the water line. While in one approach it slides a split window over the bridge pier to detect the water level, in another one it fits a linear regression line to represent the water surface. In next few sections, we describe each step in details.

\subsection{Problem Definition}
The main aim of this study is to find the difference in water levels between a reference image and new image. Let $R$ is the reference image from the AoT node where the height of water level, $h_r$ is known. For a new AoT image $I$, the goal is to find the difference in the heights of water level $\Delta h$, between images $R$ and $I$, which will subsequently compute the $h_i$, the height of water level in $I$.
\begin{equation} \label{eqn_height}
h_i = {h_r + \Delta h}
\end{equation}
Here, $h_i$ and $\Delta h$ are variables in the range of $(-\infty , +\infty)$. While $h_r$ is a positive constant, $\Delta h$ can be positive or negative.
% One image with a clear view at the bottom part of the bridge is identified to be as a reference image R. The difference calculated between the water line pre-determined for R and for the input image I is defined as the water level change in I. 

\subsection{Image pre-processing}

\begin{figure*}[!h]
    \centering
    \begin{subfigure}[h]{0.33\linewidth}
        \includegraphics[width=\linewidth]{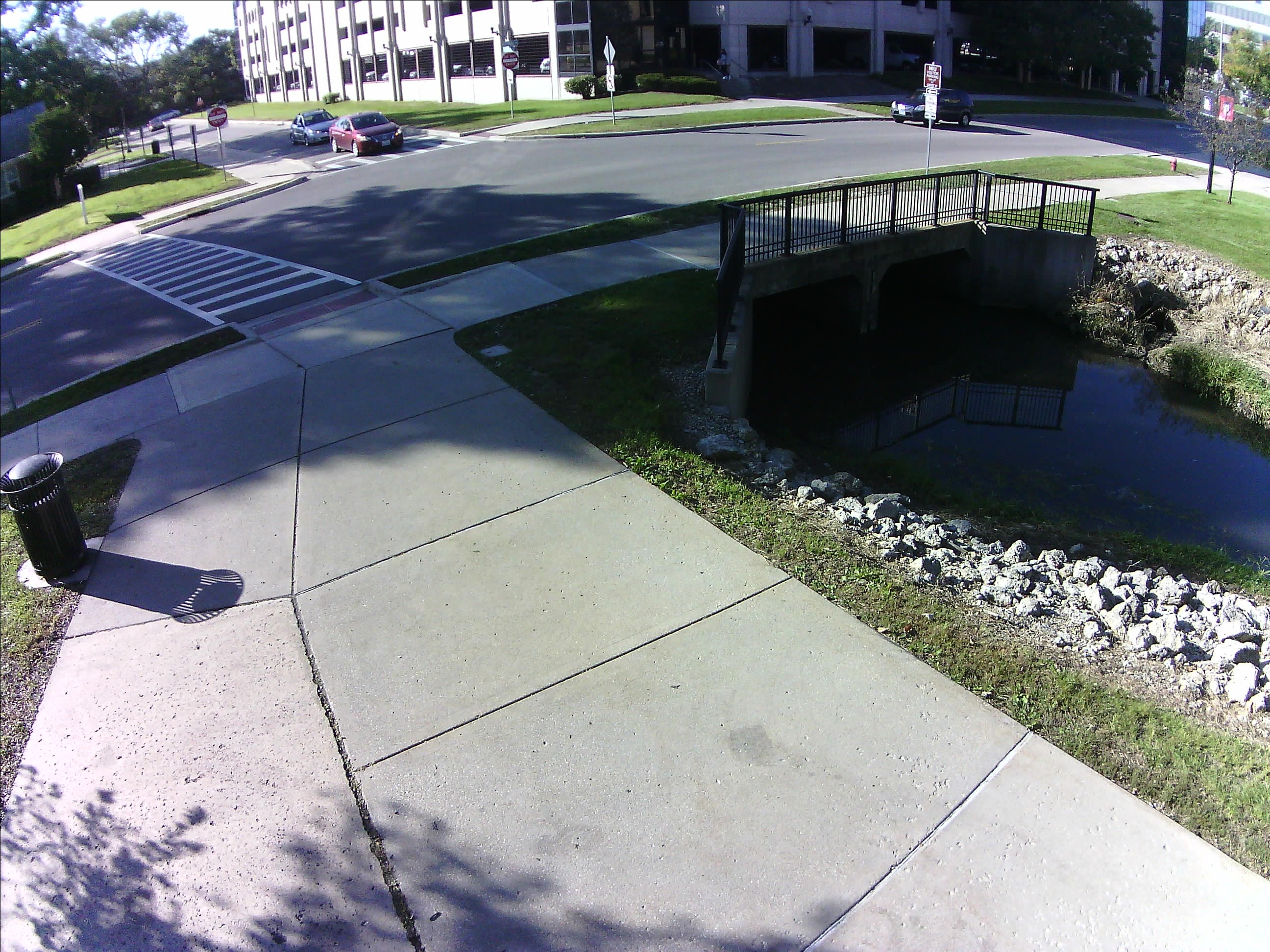}
        \caption{An original image with shadow}
    \end{subfigure}
    \begin{subfigure}[h]{0.33\linewidth}
        \includegraphics[width=\linewidth]{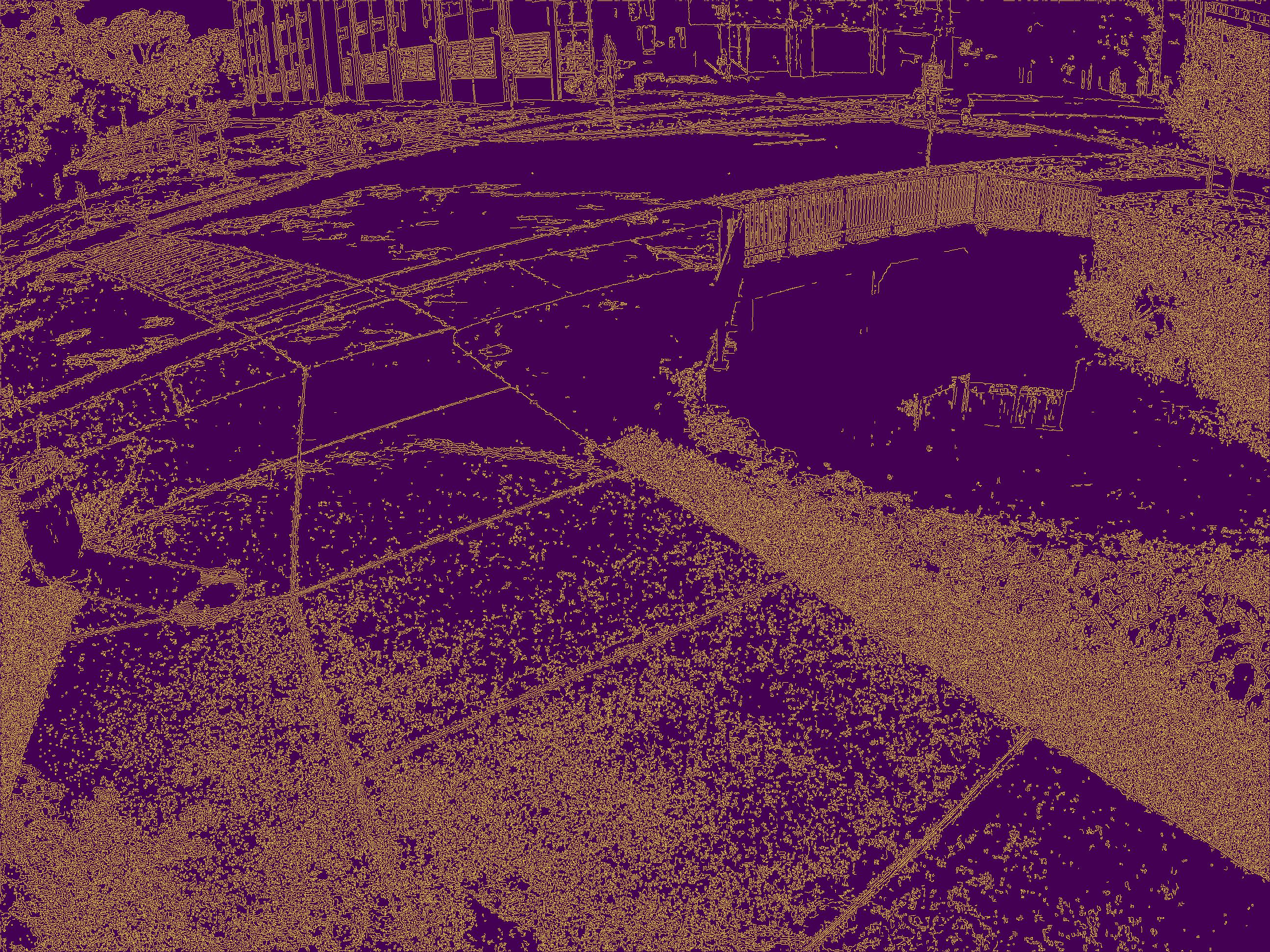}
        \caption{Edge map from Canny edge detector}
    \end{subfigure}
    \begin{subfigure}[h]{0.33\linewidth}
        \includegraphics[width=\linewidth]{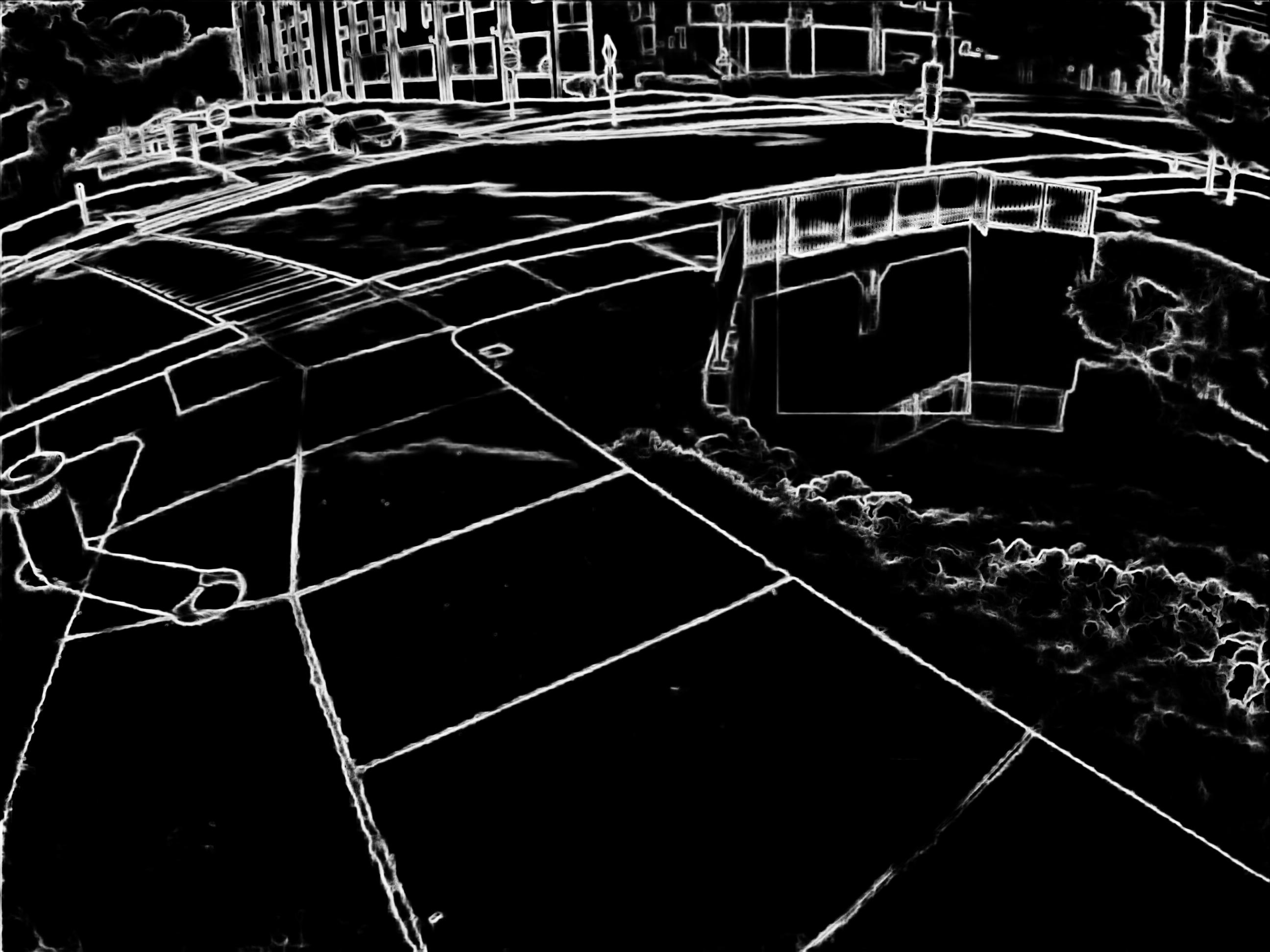}
        \caption{Edge map from HED}
    \end{subfigure}
    \caption{A comparison between Canny and HED edge detector on dark images}
    \label{fig:shadow_hed_canny}
\end{figure*}

After analyzing thousands of images collected during $3$ months, we found that the camera is fairly fixed at its position and average shift realized in vertical and horizontal direction were only $\pm 1$ and $\pm 2$ pixels respectively. We leveraged this information to segment a $400\times400$ size of region of interest from the full image (with size $2592\times1944$) that reduces the processing overhead significantly. Further, to reduce the noise in segmented ROI, image is passed through a low pass filter for smoothing by decreasing the disparity in pixel intensities among neighboring pixels. The low pass filter replaces a pixel value with the average of nearby pixels.  It retains the low frequency information while reducing the high frequency information. We used $5\times5$ surrounding pixel area for averaging.

Additionally, the system confirms the light condition of the image by randomly selecting $500$ pixels and computes their average intensity across all $3$ red, green and blue channels. It rejects the image from further processing if the average intensity across each channel is less than $30$. This process helps to avoid processing completely dark images especially during night or when a dark shadow of the tree appears on the bridge pier. A sample image is shown in Fig. \ref{fig:shadow_hed_canny} where a dark shadow of nearby tree is reducing the visibility of the creek significantly. Additionally, screening for darkness allows the system to not set a rigid time range of sunrise/ sunset to initiate processing. In this experiment, the system started processing $1-2$ hours before the sunrise and ended $1-2$ hours after the sunset. In cases where the intensity of pixels are greater than the fixed threshold but still in low range ($\leq 100$), the system enhanced the brightness of image by $50\%$ to alleviate further processing.

\subsection{Template Matching and HED}
% Using Edge detection, it is easier to find the correct zone
% Traditional Edge detector doesn't work well compared to Canny
% HED works better and is deep learning based
% Template matching is one of the simplest tool and have been used in many works
% Template matching doesn't work well in color images

% The basic idea is to use template matching to  
% 

From technical standpoint, our goal is to find the water line formed by water touching the bridge pier. if detected correctly, the mark of water line can be translated into the height of water level in the creek. Assuming water line an object, this is a classic case of object detection problem in computer vision. The current state of the art in object detection are based on trained deep neural networks to find complex objects in the image~\cite{he2017mask}\cite{tan2019efficientnet}. However, in our case object (water line) is relatively simple as it doesn't change its shape or size and only moves vertically up and down in a restricted fashion. Therefore, training and deploying a complex AI models would introduce unnecessary computing overhead. 

To avoid unnecessary overhead, we adapted a template matching algorithm \cite{brunelli2009template} (TM) that is easy to implement and works best when there is a minimal variance in the object appearance in terms of shape, size, color and its position in the image. TM algorithm has been effective in cases like pedestrian detection using wavelet templates \cite{oren1997pedestrian}, traffic light detection \cite{john2015saliency}, face detection \cite{tripathi2011face} and more. The algorithm works by sliding a predetermined template of an object over the new image. While sliding, it computes the normalized cross-correlation score $C$,  

\begin{equation}
C(x,y)=\frac{\sum_{x',y'} (T(x',y') \times I(x+x',y+y'))}{\sqrt{\sum_{x',y'}T(x',y')^2.\sum_{x',y'}I(x+x',y+y')^2}}
\label{eq:tmccornormed}
\end{equation}
between the template $T$ and the portion of the image $I$ at a position $(x,y)$ to determine the similarity, where $x$ is the row and $y$ is the column index in image $I$, and $x'$ is the row and $y'$ is the column index in template $T$. To ensure TM algorithm is effective, it is important to carefully select a template that shows minimal variance under different light and weather conditions. Initially, we tried several templates under different settings (i.e., light, weather) in color images but it didn't performed well due to high variance in RGB images induced by varying light intensity throughout the day. To reduce the variance between template and new image, we transformed the RGB image to edge map using edge detection algorithm. Edge map reduces variance by only containing the edge information in gray scale. Additionally, the edge map outlines the water line more clearly compare to RGB image. However, traditional edge detection algorithms like Sobel \cite{vincent2009sobel} and Canny \cite{canny1986computational} do not perform well in poor light conditions (as shown in Fig. \ref{fig:shadow_hed_canny} where bridge pier is covered in shadow). To improve on this, we adapted a deep neural network based holistically-nested edge detection (HED) \cite{xie2015holistically} algorithm to obtain sharp edge maps from RGB images.

The HED \cite{xie2015holistically} is an end-to-end  supervised algorithm that automatically learns rich hierarchical features important to detect consistent edges in the image. HED model is based on deep neural network trained to predict edges in an image-to-image fashion where output training image is hand annotated edge map for corresponding RGB input image. The nested layers in HED progressively refines edge maps produced at each layer that becomes more concise as it proceeds towards the output layer. Compare to Canny \cite{canny1986computational}, HED edges are consistent and stronger (comparison is shown in Fig. \ref{fig:shadow_hed_canny}). In current work, we have leveraged a pre-trained HED model trained on BSD500 dataset \cite{amfm_pami2011_bsds500} with $0.782$ F-score. Since HED is a complex neural network model, it is compute intensive as it takes $1$ second to produce edge map of a RGB image with $2592\times1944$ size using Nvidia P100 GPU. To alleviate this issue, instead of processing the whole image, only a small subsection of size $400\times400$ near ROI was used for the edge map. It helped reducing processing time for one image from $1$ second to $0.2$ second. Reduction in image size also helped in faster template search as number of comparison is reduced multiple folds. 

\begin{figure}[!ht]
    \centering
    \begin{subfigure}[h]{0.25\linewidth}
        \includegraphics[width=\linewidth]{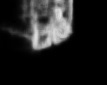}
        \caption{Template}
    \end{subfigure}
    \begin{subfigure}[h]{0.74\linewidth}
        \includegraphics[width=\linewidth]{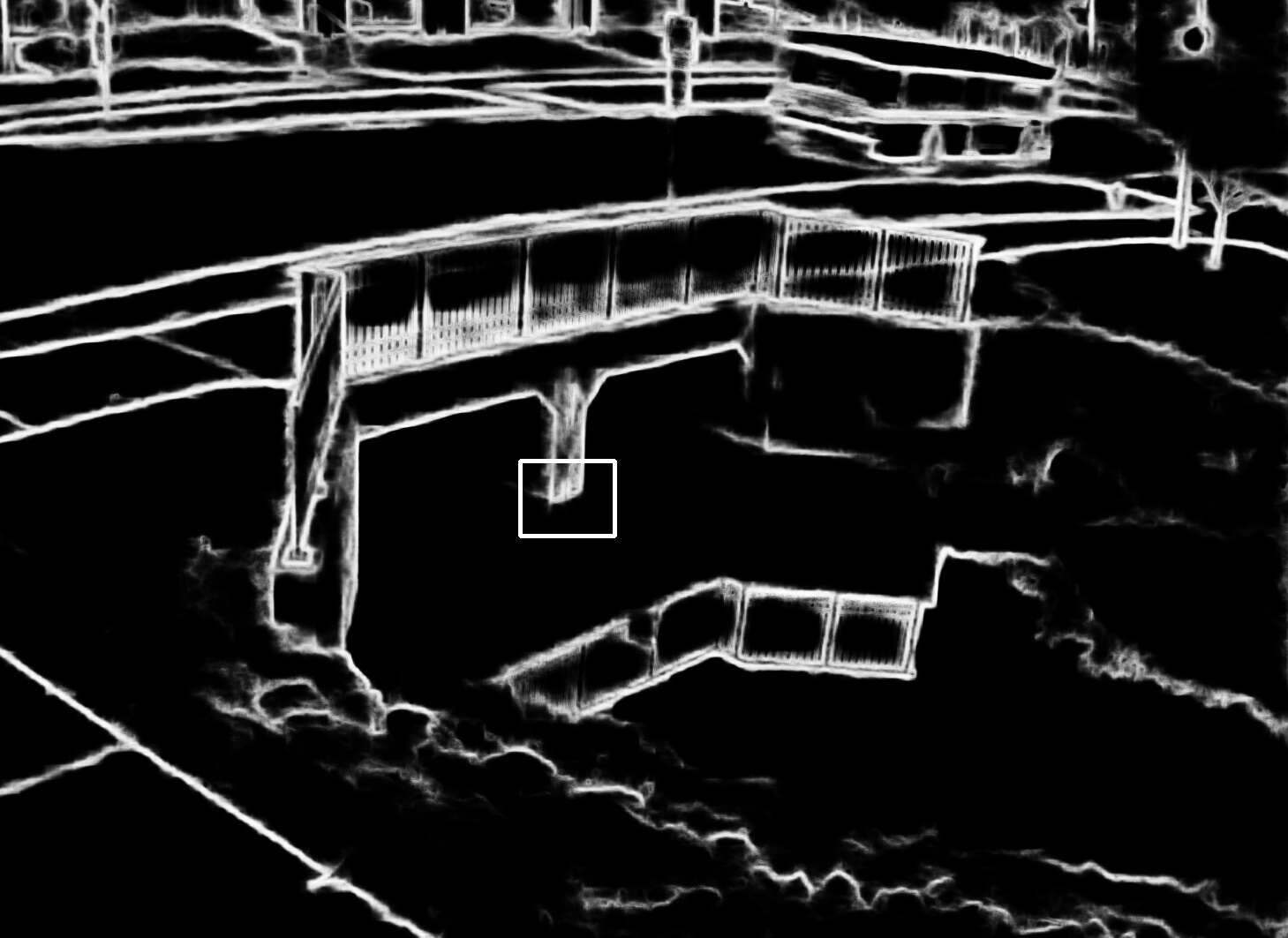}
        \caption{HED image after applying template match.}
    \end{subfigure}
    \caption{Template and output of template match.}
    \label{fig:template_and_match}
\end{figure}

As shown in Fig. \ref{fig:template_and_match}, the template is selected as a portion of edge map near the bridge pier where the water level is clearly visible and there is a high contrast in pixel values above and below the water line. In template matching, normalized cross-correlation score ranges between $0$ and $1$, where $0$ means no match and $1$ represents perfect match. The threshold score for template matching was set for $0.8$. Out of $4306$ images across $3$ months, TM algorithm found the correct match in $2218$ and $4187$ cases for RGB and edge map images, respectively. In cases where match is not found system doesn't generate any response. TM algorithm works as an effective filter for bad images. We found that in some cases images were turned upside down due to issues in camera settings that got rejected by TM algorithm due to no match found. Once the template is found in a new image, we employ two different methods to detect the water line within the matched region.

% First, the image is processed through a holistically-nested edge detection (HED)~\cite{c25} system. HED system uses a network that is a deep learning model based on fully convolutional neural networks and deep supervision. The image is passed as an input to the trimmed VGG-16 Net and multiple edge-map images are produced as output. The side outputs at each layer are further passed to deeply-supervised nets enhancing multi-scale feature learning to provide rich edges of the image. It provides state-of-the-art performance at high speed. The images obtained after applying HED are in grayscale and are used for further processes. HED is applied to the reference image. A sample image and image after applying HED can be seen in figure \ref{fig:original_and_hed}.

% To help in identifying the location in the image, where bridge and water meets, a template of the bottom part of the bridge is cropped from the reference image which is shown in figure \ref{fig:template_and_match}(a). This template is used for matching the rest of the images to detect bottom part of the bridge in the images. Template Matching is a technique to find the location of the template by searching in the original image. Template matching has been used to identify the location of eye in drivers image \cite{c26} and for road sign recognition\cite{c27}. Through template matching the coordinates of the four corners of the area in the images and the matching scores are obtained. The correctness of the match is identified based on the matching scores. In case of the images with poor lighting conditions, there will be a low matching score, and these are ignored. Figure \ref{fig:template_and_match}(b) is an example of the match found for the new image in figure \ref{fig:original_and_hed}.

\begin{figure}[!ht]
    \centering
    % \begin{subfigure}[h]{0.45\linewidth}
    %     \includegraphics[width=\linewidth]{Pictures/coordinates_template.jpg}
    %     \caption{For the template.}
    % \end{subfigure}
    \begin{subfigure}[h]{0.95\linewidth}
        \includegraphics[width=\linewidth]{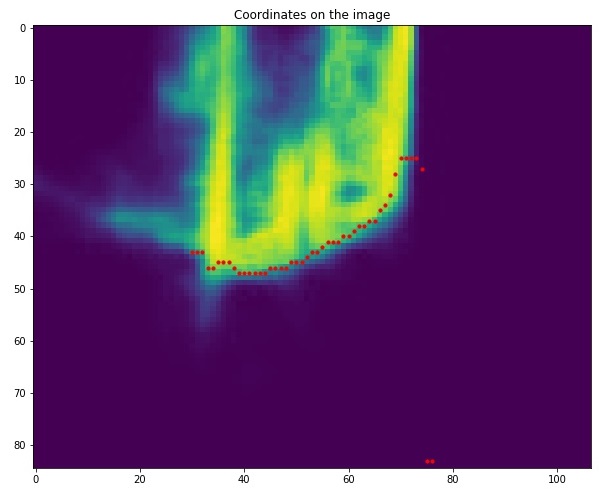}
    \end{subfigure}
    \caption{Water coordinates identified at the bridge pier.}
    \label{fig:coordinates}
\end{figure}

\subsection{Linear Regression Approach}
In our first approach, we employed the linear regression algorithm to detect the water line in the template matched region. After a successful match, the next step is to find the coordinates in the image where water surface contacts the bridge within the template matched region as shown in Fig \ref{fig:coordinates}. In edge map, pixel values ranges between $0$ and $1$, where $0$ is black and $1$ is white color. As shown in Fig. \ref{fig:coordinates}, the pixels of bridge area are bright while water are dark. Assuming $x$ and $y$ are the width and height of the image, the goal is to find the coordinate position in each column $y_i, 0 \leq i < x;$ where pixel value changes drastically. To find the precise coordinate for each column $y_i$, a $3$-pixel size window slides from bottom to top. Finally, the window with water line is selected where the first occurrence of each position in the window having greater than $70th$ percentile of that column is detected. The center position in the selected window is considered a point on water level in the column. Subsequently, coordinates are found for each column.  

Following detection of water coordinates in the image, in next step the system fits a line using linear regression algorithm. However when edge map is not sharp, water coordinates can be very noisy that makes fitting a correct line very challenging. To alleviate the noise, the coordinates are divided sequentially into $5$ equal windows and a line is fitted using linear regression for each window. The best line is picked as water line which is the most parallel to the water line in the reference image. Since the angle of water line usually remains constant, the slope of water line in reference image and new image should be similar.  

\subsection{Sum of Squared Difference Approach}
\begin{figure}[!ht]
    \centering
    \begin{subfigure}[h]{0.85\linewidth}
        \includegraphics[width=\linewidth]{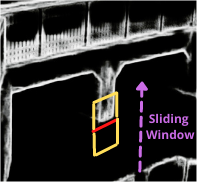}
    \end{subfigure}
    \caption{A split window sliding over the bridge pier}
    \label{fig:gradientsum}
\end{figure}

Occasionally, turbulence in the stream can make water surface noisy that may induce error in water level detection solely based on the slope comparison. To fix this, we designed a supplement method that detects the water line by computing difference in pixel intensities over a split sliding window. As shown in Fig. \ref{fig:gradientsum}, the split window slides from the bottom to top of the bridge pier and computes the sum of squared difference at each position. The window is equally split in top and bottom halves such that the top, bottom and divider lines remain parallel to the water line. As window slides from bottom to top, the sum of squared difference in pixel intensity $S(y)$ at a vertical position $y$ is computes as,

\begin{equation}
S(y)=\sum_{x',y'}{[I_u (x', y') -I_l (x', y'+h)]^2}
\label{eq:squareddifference}
\end{equation}

where, $I_u$ and $I_l$ are the upper and lower halves of the window respectively, $(x',y')$ are pixel locations in $I_u$ and $h$ is the height of each half of the window. For the vertical position $y$ where $S(y)$ score is highest, the divider line will coincide with the water level. Subsequently $y$ will be the pixel height of water level in the image.

While both of these approaches detect water level in terms of pixel distance, it can be easily converted into physical distance. To do so, we setup a reference image where pixel distance is calibrated in physical length. To compute the physical height of water level in the new image, the system first measures the difference in water level between new image and the reference image in pixel distance. Subsequently, the calibration rule from reference image is used to covert the pixel difference into physical distance. To produce the final response, we ensemble both approaches to reduce inconsistencies in the results. if the response from both approaches are within $3$ pixel distance, the final water level is computed as the mean of both responses. In any other case, both responses are rejected and the system doesn't generate any response.

\section{Performance Evaluation}

\begin{figure*}[!ht]
    \centering
    \includegraphics[width=\linewidth]{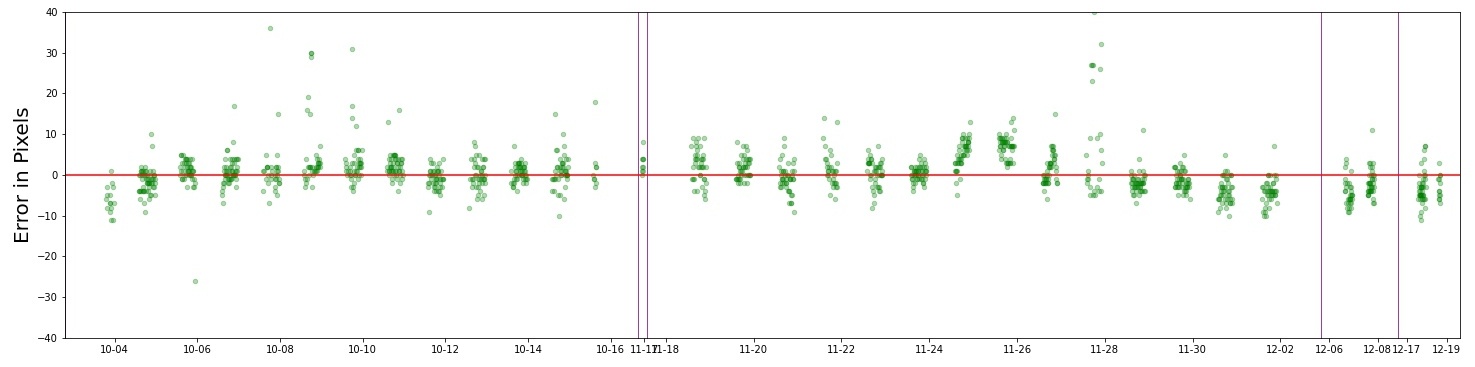}
    \caption{Error in water level detection in comparison to ground truths}
    \label{fig:error_results}
\end{figure*}

\begin{figure}[!ht]
    \centering
    \begin{subfigure}[h]{0.45\linewidth}
        \includegraphics[width=\linewidth]{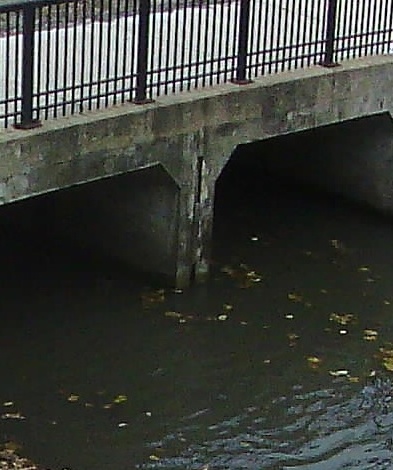}
    \end{subfigure}
    \begin{subfigure}[h]{0.45\linewidth}
        \includegraphics[width=\linewidth]{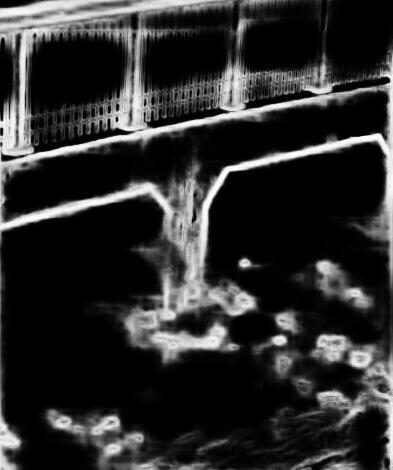}
    \end{subfigure}
    \caption{A noisy edge map due to debris flowing in the stream}
    \label{fig:debris}
\end{figure}

In this section, we evaluate the accuracy of our ensemble method against the ground truths from $4306$ images collected between $3$rd October and $18$th December in 2019 with the frequency of $1$ image in every $10$ minutes. We found that both of the proposed approaches converge in $91\%$ of their responses. Due to occasional camera and network issues, few days of images were missed in this period. Additionally, completely dark images (especially from the night) were avoided for ground truths where it was not possible to see the water level. The difference in ground truths and water level  measured by the proposed ensemble method is shown as errors in pixel distance in Fig. \ref{fig:error_results}. The majority of errors are contained within $\pm10$ pixels where $1$ pixel in image is equivalent to $1$ cm in the physical distance. On $28$th November, where error rate is more than usual, there were lot of debris flowing in the stream making edge map very noisy (as shown in Fig. \ref{fig:debris}). Also, errors are relatively high for $25$th and $26$th November because of consistent rain made the water surface uneven. Additional source of error, although minor, was induced due to human errors in collecting ground truths. Since, the bridge pier is only a small portion of the image, it is easy to under or over shoot few pixels while manually picking the exact water line. 

% The expected output of this approach shows the visualization of changes in water level on simultaneous days. The output from this method is evaluated by comparing the trends with the water level observed in south branch Kishwaukee river by United States Geographical Survey.
As the water level is recorded in continuous real values, we evaluated the proposed system with three regression-based evaluation metrics: mean absolute error (MAE), mean absolute percentage error (MAPE), and R-squared ($R^2$) as described in Equations~\ref{eq:MAE}, \ref{eq:MAPE}, and \ref{eq:RSquare}, respectively. MAE measures the average magnitude of the error without considering its direction. MAE doesn't have any bounded range since it uses the scale of ground truth data. Therefore, it can not be interpreted independently without knowing the scale of input data. In contrast, MAPE measures the percentage error in the prediction based on the computed and ground truth values. It has a bounded range from $0$ to $100$ and can be analyzed independently. Another metrics $R^2$ (coefficient of determination) is the ratio of the prediction error and the variance in the predicted values. It typically ranges between $0$ and $1$, where $R^2\leq0$ indicates that the model explains none of the variability of the data around its mean and $1$ indicates it explains all the variability.  If $h_{i}$, $\hat{h_{i}}$ and $\bar{h_{i}}$ are predicted, real, and mean predicted speeds respectively, then the equations can be defined as Equations~\ref{eq:MAE}, \ref{eq:MAPE}, and \ref{eq:RSquare}. In our case, the observed MAE, MAPE and $R^2$ scores are $4.8$, $3.1\%$ and $0.93$ respectively. In average, the height of water level (visible portion of bridge pier) are $200$ pixels while the height of image is $1944$ pixels, effectively only $10\%$ of the image is covering the bridge pier. If the bridge pier is zoomed in the image such that instead of 1 pixel representing 1 cm of physical distance, it represents $1$ mm, the error rate can be improved further in terms of physical height of water level.

We also compared our results against USGS data. The water in the creek branched from the XXXXX river in the city of XXXXX. The USGS records the water level of the river in XXXXX in every $15$ minutes. The dataset is publicly available on the USGS website. Although the flow of water is controlled between the USGS station in the river and the creek, we observed a strong correlation between the water levels detected by our approach and reported by USGS. Between $3$rd October and $18$th December in $2019$, the $R^2$ score between both readings is $0.88$. Encouraged by this result, we believe that our solution can supplement the USGS monitoring with high sampling frequency.

%%Metrics - MAE, MAPE, RSquare
\vspace{-0.5em}
\begin{equation}\label{eq:MAE}
    MAE=\dfrac{1}{n}\sum_{i=1}^{N}|h_{i}-\hat{h_{i}}|
\end{equation}
\begin{equation}\label{eq:MAPE}
    MAPE=\dfrac{1}{n}\dfrac{\sum_{i=1}^{N}|h_{i}-\hat{h_{i}}|}{h_{i}}
\end{equation}
\begin{equation}\label{eq:RSquare}
    R^2=1-\dfrac{\sum_{i=1}^{N}(h_{i} - \hat{h_{i}})^2}{\sum_{i=1}^{N}(h_{i} - \bar{h_{i}})^2}
\end{equation}
\vspace{-1.75em}

\section{Discussion}
The proposed solution works best when the edge map is sharp and water surface is clearly outlined. In cases when edge map is noisy, the system either doesn't produce a response or have high variance in their measurements within a short time frame. Although both scenarios are bad, relatively, providing wrong information is worse than giving no information at all. To mitigate the high variance issue, the system must also provide their confidence in their measurement. To do so, variance can be mapped to error rate using historical data and subsequently into confidence intervals. So, when variance is high, confidence should be low and vice-versa. On the other hand, the system generates no response either due to template matching algorithm is not able to find the bridge pier or both approaches don't converge on single response. In the latter case, we can use a weighting system to give more weight to one approach over other based on their historical performance. In critical scenario, when it is raining heavily and system has low confidence in measurement it can simply send the raw image of creek to the user. While in the former case, we will explore a lightweight AI based object detection approach that can outperform template matching without compromising the efficiency too much. All of the above discussed problems are open ended which we intend to explore in near future.
\section{Conclusion}
In this work, we have proposed an ensemble method based on edge map to detect the water level in the stream. The proposed method combines two unique approaches to identify the water surface in the edge map. Since, the accuracy of our method depends on the quality of input data, we have employed state-of-the-art pre-trained HED model to generate the edge map. While the first approach uses the linear regression algorithm to fit the line in water coordinates, second approach computes the sum of squared difference between a split sliding window to detect the water line. In evaluation, our proposed solution has achieved $4.8$, $3.1\%$ and $0.92$ scores for MAE, MAPE and $R^2$ metrics, respectively. The current solution employs lightweight image processing algorithms and template matching filters to reject noisy image and inconsistent results to produce accurate, fast and trustworthy responses. Unlike other vision based solution, the current approach doesn't require any additional infrastructure like staff gauge installed in the stream. Also, it doesn't require to train a complex deep neural networks model to measure the accurate water level. Our solution can be easily adopted to another location by pointing the camera towards the bridge pier. Due to low cost implementation, multiple cameras can be installed in a local region to crowd source the water level information to collect reliable results. In future, we would like to improve our solution by reducing the noise from the edge map induced by flowing debris or heavy rain. Additionally, we would want to analyze the system performance for longer period and in multiple locations.

\bibliographystyle{ACM-Reference-Format}
\bibliography{bibliography} 

\end{document}